\documentclass[times,twocolumn,final]{elsarticle}
\usepackage{graphicx}              
\DeclareGraphicsExtensions{.pdf,.jpg,.jpeg,.png} % for pdflatex we expect .pdf, .png, or .jpg files

\usepackage{cag}
\usepackage{framed,multirow}
\usepackage{microtype}
\usepackage{booktabs}
\usepackage{float}
\usepackage{amsmath}
\usepackage{subfigure}
\usepackage[table,xcdraw]{xcolor}
\usepackage{amssymb}
\usepackage{latexsym}
\usepackage{enumitem}
\usepackage{cuted}
\usepackage{ccicons}
\usepackage{pifont}
\usepackage{tabularx}
\usepackage{multirow}
\usepackage{graphicx}
\usepackage{caption}
\usepackage{stfloats}
\usepackage{url}
\usepackage[table]{xcolor}

\usepackage{hyperref}

\usepackage{comment}

\journal{Computers \& Graphics}

\renewcommand*{\appendixautorefname}{\kern-3pt} % fix the weird appendix naming behavior

\begin{document}

\begin{frontmatter}
\title{Evaluating Graphical Perception Capabilities of Vision Transformers\vspace{-1pt}}%

    \author[1]{Poonam \snm{Poonam}\fnref{fn1}\corref{cor1}}
    \author{Pere-Pau \snm{Vázquez}\fnref{fn2}}
    \author{Timo \snm{Ropinski}\fnref{fn1}}
    \fntext[fn1]{Viscom Group, Institute of Media Informatics, Ulm University}
    \fntext[fn2]{ViRVIG Group, Universitat Politècnica de Catalunya - BarcelonaTech}

\cortext[cor1]{Corresponding author: }
    \emailauthor{{poonam.poonam, timo.ropinski}@uni-ulm.de}{Poonam Poonam, Timo Ropinski}
    \emailauthor{pere.pau.vazquez@upc.edu}{Pere-Pau Vázquez}
    \address[1]{Ulm University, Ulm, 89081, Germany}

\received{17th Sept 2025}

\begin{abstract}
    Vision Transformers (ViTs) have emerged as a powerful alternative to convolutional neural networks (CNNs) in a variety of image-based tasks. While CNNs have previously been evaluated for their ability to perform graphical perception tasks, which are essential for interpreting visualizations, the perceptual capabilities of ViTs remain largely unexplored. In this work, we investigate the performance of ViTs in elementary visual judgment tasks inspired by Cleveland and McGill’s foundational studies, which quantified the accuracy of human perception across different visual encodings. Inspired by their study, we benchmark ViTs against CNNs and human participants in a series of controlled graphical perception tasks. Our results reveal that, although ViTs demonstrate strong performance in general vision tasks, their alignment with human-like graphical perception in the visualization domain is limited. This study highlights key perceptual gaps and points to important considerations for the application of ViTs in visualization systems and graphical perceptual modeling.
\end{abstract}

\end{frontmatter}
%-----------------------------------------------------------
%-----------------------------------------------------------
\section{Introduction}
% role of transoformers in vision and data visualization
Vision-based image analysis has undergone a significant transformation with the emergence of vision transformers (ViTs)~\cite{vaswani2017attention}. Transformer architectures have been successfully adapted to visual data, demonstrating superior ability to capture complex spatial patterns as compared to traditional convolutional neural networks (CNNs)~\cite{dosovitskiy2020image, chen2023pali, liu2022matcha, masry2023unichart, Cheng_2023_ICCV, han2023chartllama}.
This transformation also influences areas of computer graphics, where vision models increasingly support visualization systems, chart interpretation, and graphics perception tasks.
CNNs inspired by the hierarchical organization of the human visual cortex~\cite{hubel1959receptive}, emphasize local spatial features through layered convolutions~\cite{itti2001computational, kim2016convolutional}. However, their reliance on local receptive fields limits their ability to integrate global context across an entire image~\cite{Jacobsen_2016_CVPR}. In contrast, ViTs utilize self-attention mechanisms that dynamically weight relationships across the entire image, enabling simultaneous modeling of local details and long-range dependencies~\cite{https://doi.org/10.1002/wcs.1570}. This integrated processing aligns with cognitive attention mechanisms~\cite{article} and gives ViTs a distinct advantage in interpreting complex spatial relationships, as global contextual understanding is particularly critical in data visualization analysis, where meaningful interpretation depends on relating multiple visual elements—such as data points and legends—in context~\cite{lal2023lineformer,yan2023context}. 
Consequently, ViTs offer a promising framework for advancing computational approaches to chart analysis, computer graphics, and visualization comprehension. 

However, despite their strengths in capturing complex spatial patterns, it remains unclear how well ViTs align with human visual perception in fundamental low-level visual tasks-an alignment that is essential when applying these models to data visualization. Accurate perception of elementary visual encodings such as position, length, angle, and area forms the basis of graphical understanding, as extensively characterized by Cleveland and McGill’s foundational experiments~\cite{cleveland1984graphical}. Their work established a hierarchy of human perceptual accuracy that continues to inform visualization design and evaluation~\cite{franconeri2021science}.

Building upon this framework, our study investigates the ability of ViTs to replicate human performance on core perceptual tasks. We compare the performance of ViTs to that of CNNs and humans on fundamental graphical perception tasks. 
By situating our study within the framework of perception-driven visualization research, we contribute to the graphics community’s efforts to evaluate machine models not only by task performance, but also by perceptual fidelity to human viewers.
Since transformers represent the current state-of-the-art in chart analysis~\cite{carbune2024chart}, understanding their ability to replicate human perceptual accuracy at this basic level is essential for determining their effectiveness in visualization-related applications, ranging from automated chart interpretation to perceptually informed design.

To explore how ViTs perform on low-level visual tasks, we conducted perceptual evaluations of three representative ViT architectures: the vanilla Vision Transformer (vViT), the Convolutional Vision Transformer (CvT), and the Shifted Window Transformer (Swin). These models were selected to reflect architectural diversity within the ViT family, capturing key differences in tokenization strategies, convolutional integration, and spatial hierarchy. 
Hence, we make the following contributions in this paper:
\begin{itemize}
    \item We evaluate the performance of three canonical ViT architectures on low-level visual tasks, replicating and extending the perception experiments conducted by Cleveland and McGill~\cite{cleveland1984graphical}.
     \item We compare ViTs performance to that of CNNs and human observers, and discuss the implications for perceptual alignment in visualization systems.
\end{itemize}

These findings shed light on the evolving role of ViTs in the data visualization research agenda—particularly in understanding, generating, and redesigning data visualizations. In addition, we reflect on limitations and outline potential directions for future research. 

\section{Related Work}
\noindent\textbf{Evaluating human perception.} The process by which humans interpret visual information has been a central focus in various fields including information visualization, data science, computer vision, and human-machine interaction. One of such insights dates back to the 50s, with work done by Hubel and Wiesel~\cite{hubel1959receptive}, whose seminal work established a foundational understanding of how visual information is processed in the brain, from the detection of simple features such as edges, to the recognition of complex shapes and objects. Such understanding of how people perceive and interpret visual representations is crucial for designing effective visualizations that facilitate accurate comprehension. In their influential work, Cleveland and McGill conducted systematic experiments to evaluate the effectiveness of different graphical representations~\cite{cleveland1984graphical}, by tasking humans to solve several low-level visual tasks. Based on their study, they shed light on the hierarchical nature of graphical effectiveness. 
This perceptual ranking has since been adopted in many visualization design systems~\cite{mackinlay1986automating} and empirically validated in follow-up experiments using modern platforms such as Mechanical Turk~\cite{heer2010crowdsourcing, talbot2014four}, supporting its enduring relevance to computer graphics and visual design.
Building on these foundational studies, a recent survey on information visualization~\cite{liu2014survey} reviewed advances and challenges in the field, offering a comprehensive overview of how perceptual and cognitive factors continue to shape visualization research.

Extending beyond fundamental perceptual skills, later studies examined how users cognitively interpret and reason with unfamiliar or complex visualizations.
These investigations examine not only accuracy in perception but also the cognitive strategies, abilities, and evaluation methods involved in comprehending graphical information.
Kong et al.~\cite{kong2017internal} explored how presenters and viewers prefer to highlight or cue parts of a chart to draw attention.
Boy et al.~\cite{6875906} studied how individuals interpret graphs and assessed their ability to understand and extract meaning from graphical information rather than on basic low-level perceptual tasks. 
Lee et al.~\cite{7192668} explored how individuals derive meaning from unfamiliar graphical data and use their cognitive abilities to interpret them.
More recently, a visually-supported topic modeling approach for spatio–temporal events~\cite{MOUSSAVI2025104245} showed how integrated visual analytics can help users identify behavioral patterns, underscoring the role of perceptual and cognitive support in reasoning with complex data.
Zhu et al.~\cite{zhu2025esiqa} created the first database for egocentric spatial images and proposed ESIQAnet, a model that predicts perceptual quality across 2D and 3D display modes.
Wang et al.~\cite{wang2023perceptual} further demonstrated that size discrimination thresholds vary between AR and VR, offering insights into how visual environments influence perceptual sensitivity.
Sterzik et al.~\cite{sterzik2023enhancing} empirically evaluated line-style variables for uncertainty visualization in molecular graphics, highlighting how encoding choices like width significantly impact perceptual accuracy.
Börner et al.~\cite{doi:10.1177/1473871615594652} examined how individuals approach interpreting visualizations, including their descriptive strategies and tendencies in visual encoding preferences. 
Alves et al.~\cite{alves2023exploring} showed that individual differences such as conscientiousness can shape how users engage with visualizations and revise decisions, reinforcing the need to consider cognitive variability in perception-focused evaluations.

\noindent\textbf{Evaluating computational models.} CNNs~\cite{726791} were among the first computational models that were able to showcase remarkable capabilities in image recognition~\cite{NIPS2012_c399862d, He_2016_CVPR, pmlr-v97-tan19a}, object detection~\cite{Redmon_2016_CVPR, ren2015faster, He_2017_ICCV}, and semantic segmentation~\cite{10.1007/978-3-319-24574-4_28, Chen_2018_ECCV}. However, nowadays ViTs outperform CNNs across these and other tasks~\cite{dosovitskiy2020image, liu2021swin, wu2021cvt, Wang_2021_ICCV, pmlr-v139-touvron21a}. Naturally, the ability of these computational methods, in extracting low-level features from visual data with human perception, has become increasingly relevant to the  visualization research community. 
This intersection between vision models and visualization has also been explored within the computer graphics community, where efforts have been made to model human perception~\cite{szafir2017modeling, lu2021modeling} and test perceptual accuracy in chart comprehension using machine learning models~\cite{yang2023can}. 
Zhou et al.~\cite{wang2021survey} surveyed how machine learning techniques have been applied to data visualization (ML4VIS), outlining advances and future challenges in this emerging area.
Garcia et al.~\cite{GARCIA201830} surveyed visual analytics methods for deep learning, proposing a taxonomy for architecture understanding, training analysis, and feature interpretation. These studies highlight how perception-driven evaluation contributes to visual encoding and system design. 
Munz et al.~\cite{MUNZ202245} developed a visual analytics system for neural machine translation that combines attention and beam search visualizations with interactive correction to enhance translation quality and domain adaptation. Along similar lines, Hoedecke et al.~\cite{ScrutinAI} presented ScrutinAI, a visual analytics tool for semantically assessing object detection models, further demonstrating how interactive visualization can support the evaluation of computational perception.
In a related direction, GEMvis~\cite{chen2022gemvis} introduced a visual analysis method for comparing and refining graph embedding models, further illustrating how visualization can support the evaluation and improvement of complex machine learning representations. Complementing these model-focused systems, VizAssist~\cite{bouali2016vizassist} introduced an interactive assistant for visual data mining, demonstrating how guidance and interaction can enhance users’ ability to explore, interpret, and act on complex data. Graphics and visualization research has also examined perceptual fidelity, addressing rendering error minimization, spectral image compression, texture blending, and perceptually guided illumination~\cite{wolfe2025importance,fichet2025compression,wronski2025gpu,majercik2020scaling}.

Numerous research studies have investigated the effectiveness of CNNs in this context. For instance, Haehn et al.~\cite{haehn2018evaluating} conducted a comprehensive analysis, evaluating the capabilities of CNN models on perceptual tasks relevant to visualization. Their study provided valuable insight into the nuanced interplay between CNN architectures and the complexities of visual data interpretation. Funke et al.~\cite{10.1167/jov.21.3.16} addressed the complexities of comparing human and computational model's visual perception while also outlining strategies to overcome potential challenges in experimental design and inference. Kim et al.~\cite{kim2019neural} further gained insights into the similarities and disparities between neural networks' visual perception and human vision. By investigating concepts such as the law of closure, they demonstrated a correlation between a network's capacity to extract features and its ability to generalize. Focusing on applied visualization tasks such as scatter plot interpretation, Yang et al.~\cite{10.1145/3544548.3581111} investigated how CNNs interpret the visual features and how they can aid visualization perception. Liu et al.~\cite{liu2022deplot, liu2022matcha} studied how CNNs extract structural information from simple visualizations, and used it to classify different types of charts. While there are many other works following a similar line of research~\cite{9174891, luo2021chartocr, davila2019icdar, davila2022icpr, hartwig2025hpscan, fan2018fast, hartwig2022learning, trautner2021line}, discussing all would be beyond the scope of this paper. In contrast to these methods, other works emphasized the utilization of CNNs for extracting data, interpreting visualizations and further redesigning visualizations and answering visual questions effectively~\cite{dai2018chart, luo2021chartocr, poco2017reverse, zhou2021reverse, sedlmair2015data}. 
Beyond classification or extraction, some works have also explored the modeling of perceptual similarity~\cite{funke2021five} and visualization literacy in computational agents~\cite{dong2025probing}, signaling a broader move toward evaluating the perceptual fidelity of AI in visualization.
Similarly, Le et al.~\cite{le2023textanimar} introduced TextANIMAR, a text-based 3D animal retrieval framework that leverages fine-grained semantic features, highlighting the importance of aligning machine representations with human perceptual distinctions in complex visual domains.
Recent surveys and applications further emphasize the importance of perceptual alignment and interpretability in computer vision and visualization, spanning comparative analyses of CNNs and ViTs, perceptually motivated architectures such as capsule networks, trustworthy model–human interaction frameworks, clarity in result presentation, and generative perceptual systems~\cite{nafea2024short,mahdi2024object,noori2023towards,shin2024revolutionizing,mohialden2024agent}.
 
Recent research has increasingly applied ViTs to chart-related tasks such as classification, data extraction, and element detection. Dhote et al.\cite{davila2019icdar} compared ViTs and CNNs for chart classification, while Shivasankaran et al.\cite{hassan2023lineex} and Lal et al.\cite{lal2023lineformer} used ViTs for extracting data from line charts. Yan et al.\cite{yan2023context} combined CNNs and ViTs to improve element detection and chart redesign generalization.
Xue et al.~\cite{xue2023chartdetr} introduced a unified framework employing ViTs to detect data across different types of charts (bar, line, pie), facilitating the extraction of information. Recently, with the emergence of vision language models, researchers such as Chen et al.~\cite{chen2023pali} and Tang et al.~\cite{Tang_2023_CVPR} have also employed private proprietary projects  to interpret charts, enabling them to answer certain visual questions effectively. More recently, Dong and Crisan~\cite{dong2025probing}  evaluated vision-language models (VLMs) on data visualization tasks, highlighting their strengths and limitations in interpreting chart types and encodings, and emphasizing the need for further research on their perceptual alignment with humans.

Prior work has largely applied ViTs to chart-level interpretation tasks, such as classification or element detection. However, to the best of our knowledge, no prior work has thoroughly examined ViTs' capabilities on low-level visual tasks. 
This is especially relevant to the field of computer graphics, where recent studies have started to examine how deep models align with perceptual hierarchies~\cite{haehn2018evaluating,yang2023can}, and how vision-based agents can support visualization analysis and interpretation tasks~\cite{zhao2021chartstory}.
To fill this gap in the literature, we took inspiration by Haehn et al.~\cite{haehn2018evaluating}, and extended their work by assessing how ViTs perform on the same class of tasks.
In this study, we focus on self-trained ViT models rather than large foundation models. This choice offers greater control over model architecture, training data, and task-specific fine-tuning—factors that are essential when evaluating perceptual alignment on narrowly defined, low-level visualization tasks.

\section{Methods}
A central challenge in using neural models for data visualization lies in understanding how closely their perceptual processes align with those of humans. While many recent studies evaluate these models on complex interpretive tasks, such comparisons often obscure whether they can perceive the fundamental visual encodings—such as position, length, angle, and area—that underpin graphical comprehension. To address this, we evaluate ViTs on low-level perceptual tasks, drawing on the framework established by Cleveland and McGill~\cite{cleveland1984graphical} and extended by Haehn et al.~\cite{haehn2018evaluating}.
These low-level tasks involve the direct interpretation of visual primitives—marks, positions, shapes, or ratios—that support higher-level understanding in visualizations. Assessing ViTs on such encodings allows us to examine their alignment with foundational perceptual processes, which are critical for accurate and efficient chart interpretation.

In the following subsections, we first examine the key aspects of the tested low-level visual tasks, before describing the characteristics of the visual stimuli employed throughout our
investigations. Finally, we introduce the evaluated ViTs, and provide information about their training process.

\subsection{Low-level visual tasks}{\label{low}}
As previously mentioned, low-level visual tasks were extensively studied in the seminal work of Cleveland and McGill~\cite{cleveland1984graphical}. They thoroughly measured and ranked the accuracy of human perception in these fundamental tasks, establishing a perceptual hierarchy that remains a cornerstone of data visualization design and evaluation. These tasks revolve around interpreting basic visual encodings, the essential building blocks of data visualizations, and include the following nine elementary perceptual encodings (see Fig.~\ref{figure_ele}).

\noindent\textbf{Position along common scale.} Determining the location of data points or objects along a shared axis, facilitating comparisons or correlations.

\noindent\textbf{Position along non-aligned scales.} Assessing the relative positioning of elements across different scales or axes, aiding in understanding multidimensional relationships.

\noindent\textbf{Length.} Estimating the magnitude or extent of graphical elements, such as bars or lines, to gauge quantitative values.

\noindent\textbf{Direction.} Recognizing the orientation or trajectory of visual components, influencing interpretations of trends or patterns.

\noindent\textbf{Angle.} Assessing the angular relationships between elements, often crucial in interpreting pie charts or radial diagrams.

\noindent\textbf{Area.} Evaluating the spatial extent or proportionality of enclosed regions, guiding perceptions of magnitude or distribution.

\noindent\textbf{Volume.} Perceiving the three-dimensional space occupied by graphical objects, relevant in visualizations involving depth or volumetric data.

\noindent\textbf{Curvature.} Discerning the degree of curvature or bending within graphical representations, influencing perceptions of smoothness or complexity.

\noindent\textbf{Shading.} Interpreting variations in light and darkness to infer depth, texture, or emphasis within visualizations.

For our study, drawing inspiration from the experiment of Cleveland and McGill~\cite{cleveland1984graphical}, we explore various tasks emerging from these basic encodings, including position-length, position-angle, bars, and framed rectangles. As another relevant task, we also investigate point cloud estimation, which we have borrowed from Haehn et al.~\cite{haehn2018evaluating}. Thus, the investigated tasks are defined as follows:

\noindent\textbf{Position-angle.} 
In this task, position and angle ratios need to be estimated, which can be achieved by conducting comparisons using both bar charts and pie charts (see Fig. \ref{figure_pa}). The task is to estimate the ratio between the four smaller bars or sectors and the largest bar or sector, which is clearly visible and marked.

\noindent\textbf{Position-length.}
This task relies on position and length comprehension across various designs of grouped and divided bar charts. Despite both types of charts presenting identical data, they entail distinct elementary perceptual tasks. Grouped bar charts demand assessments of positions along a shared scale, whereas divided bar charts add the complexity of length evaluations. Fig.~\ref{figure_pl} depicts the different types, whereby types 1, 2, and 3 primarily revolve around the evaluation of positions along a common scale, and types 4 and 5 entail the assessment of length measurements. We define the range of difficulty in ascending order for these types, as it was also classified by Cleveland and McGill~\cite{cleveland1984graphical}.

\noindent\textbf{Bars and framed rectangles.} In this investigation, we assess bars and framed rectangles (see Fig.~\ref{figure_bf}) to compare judgments of length and position along non-aligned scales. Visual framing aids estimation by providing reference boundaries that improve judgments of maximum bar length. 

\noindent\textbf{Point cloud.} This task involves exploring a random 2D point cloud by considering Weber's Law~\cite{householder1940weber}. The task requires assessing the number of dots (10,100,1000) in a scatter plot as depicted in Fig.~\ref{figure_wb}, and is subject to the just noticeable difference problem~\cite{stern2010just}. 

Through exhaustive analysis of these low-level visual tasks, we gain deeper insights into how humans, CNNs and ViTs process visual information,enabling direct comparisons of their  alignment on low-level perceptual tasks.

\subsection{Data}
\noindent\textbf{Visual stimuli and labels.}
For our study on low-level visual tasks, we reproduced the Cleveland and McGill stimuli using the same process as used by Haehn et al.~\cite{haehn2018evaluating}. In their paper, they created stimuli as 100$\times$100 binary images and we followed this procedure in order to make our results comparable. We also developed a parameterized stimuli generator for each elementary task, with the number of possible parameter values differing per experiment. To ensure consistent value scaling, we normalize the generated image's pixel values to the range of  $-0.5$ to $0.5 $ before using the images. We additionally add random noise to each pixel to avoid memorization of the images by the networks. There is a ground truth label associated to the image which represents the parameter used during the generation of the image. The labels are also scaled in the range of $0.0$ to $0.1$ and normalized to the maximum and minimum value as done by Haehn et al.~\cite{haehn2018evaluating}. However, differing from their procedure, we resize the generated images to the standard input size of $224 \times 224$ pixels, as required by the ViTs. Fig.~\ref{figure_ele} shows examples of these generated visual stimuli.

\noindent\textbf{Data Splitting.}
Following Haehn et al.~\cite{haehn2018evaluating} we use 100k images per task. We partition this dataset into the training, validation, and test sets in a ratio equal to $0.6:0.2:0.2$. To create these datasets, we generate stimuli from random parameters and add them to the sets until the target number is reached, while maintaining distinct (random) parameter spaces for each set to ensure that there is no leakage between training and validation/testing.

\subsection{Vision Transformer Architectures}
In this study, we evaluate three ViT architectures that represent distinct design variations within the broader ViT family. Our goal is to assess how different architectural choices affect performance on low-level perceptual tasks. To this end, we selected: (1) the vanilla Vision Transformer (vViT), which follows the original transformer design without additional inductive biases; (2) the Convolutional Vision Transformer (CvT), which introduces convolutional layers to enhance local feature extraction; and (3) the Swin Transformer (Swin), which incorporates hierarchical representation learning and localized attention through shifted windows.

These models span a range of architectural strategies—from purely token-based global attention to hybrid and spatially aware designs—and thus offer a diverse basis for analyzing how ViTs process basic visual encodings relevant to data visualization.
\subsubsection{Vanilla Vision Transformer}
Vanilla Vision Transformer (vViT)~\cite{dosovitskiy2020image} uses the transformer architecture that was initially built for sequential language models. Unlike CNNs where images are used as feature maps, in vViT images are divided into sequential tokens. Therefore, as a very first step, images are split into patches of fixed size and then flattened. Each flattened image patch is mapped to a lower-dimensional trainable linear projection. The output of this projection is called patch embedding. Learnable embeddings are prepended to these sequential patch embeddings as illustrated in Fig.~\ref{fig_cvt}. After adding an extra positional embedding, these sequence vectors are fed into the transformer encoder~\cite{vaswani2017attention}. A feed forward network multi-layer perceptron (MLP) head stacked on top of the transformer at the position of the extra learnable embedding, that was added to the sequence, is used to classify the images.

A transformer encoder block consists of  a multi head self-attention layer (MSA),  an MLP layer, and layer norm (LN)~\cite{ba2016layer}. MSA is a kind of self-attention in which $k$ self-attention operations run in parallel, and are refereed to as heads. This layer concatenates all the attention outputs linearly to the right dimensions. The $k$ attention heads help train local and global dependencies in an image. The MLP layer contains a double-layer with a Gaussian error linear unit (GELU)~\cite{hendrycks2016gaussian} non-linearity. LN is used before each block to help to improve the training time and overall performance. After each block, residual connections are applied which improve the gradient flow without passing through non-linear activations. The trainable parameters for vViT amount to approximately 5.67 million.

\subsubsection{Convolutional Vision Transformer} 
\begin{figure*}[!t]
  \centering 
  \subfigure[vViT architecture]{
    \includegraphics[width=0.7\textwidth]{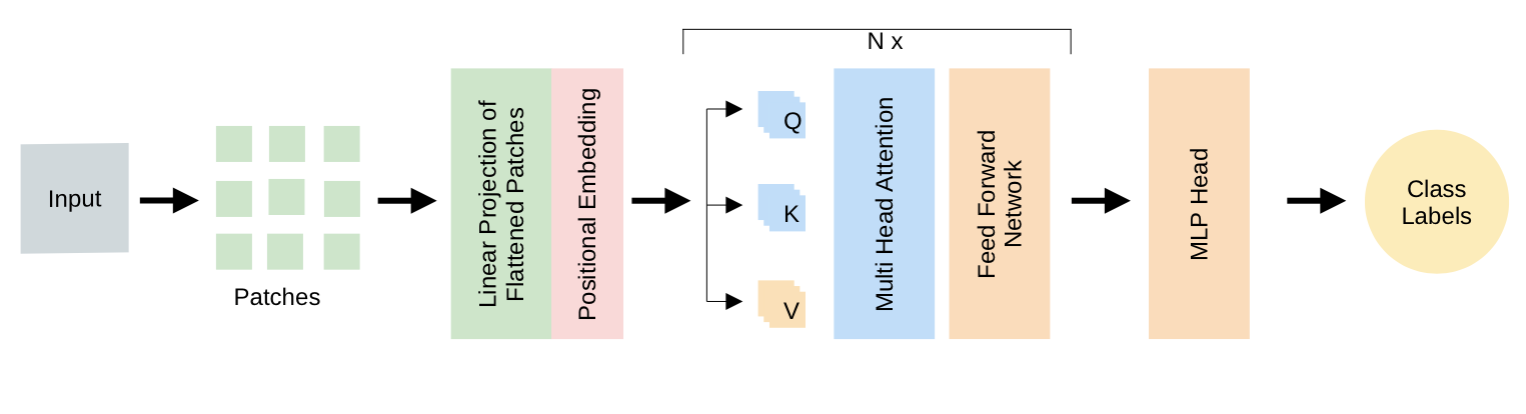}
  }\\[1ex]
  \subfigure[CvT architecture]{
    \includegraphics[width=0.7\textwidth]{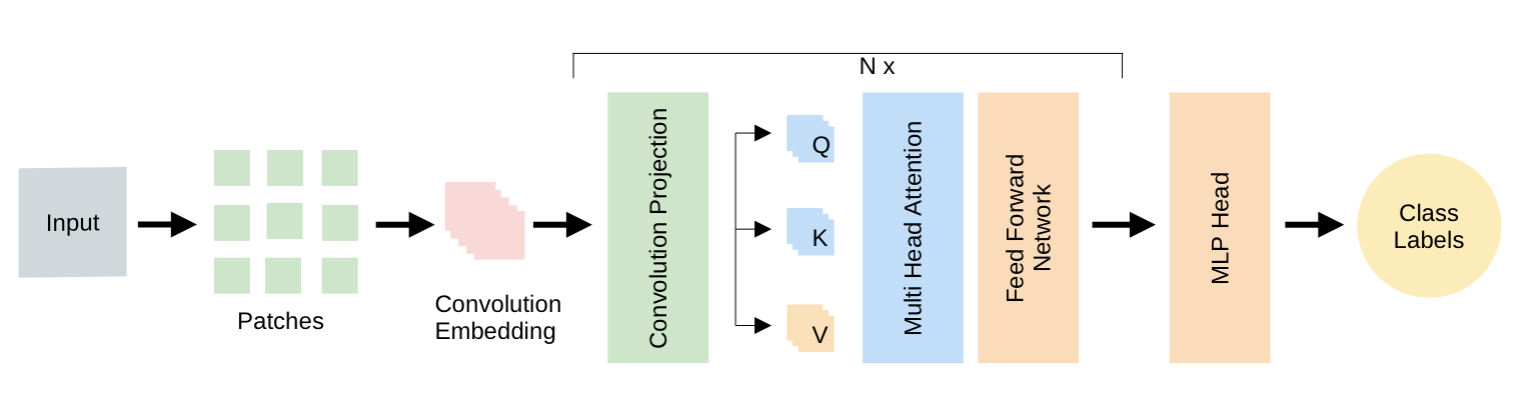}
  }\\[1ex]
  \subfigure[Swin architecture]{
    \includegraphics[width=0.7\textwidth]{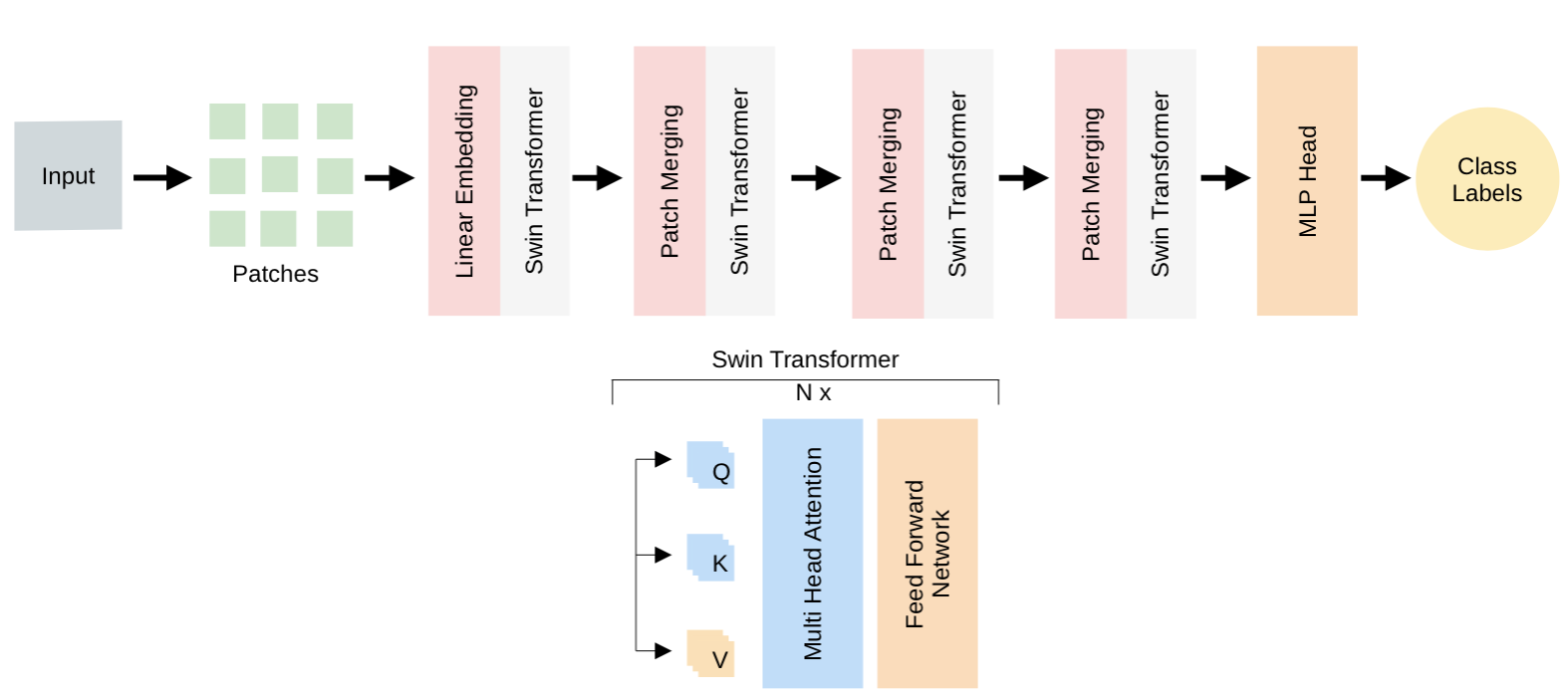}
  }
  \caption{\label{fig_cvt}%
  Illustration of the tested ViT architectures: vViT (a), CvT (b), and Swin (c). Input image patches (\emph{green}) can be projected linearly (vViT) or through convolution (CvT). Transformer blocks with self-attention (\emph{blue}) process the projected patches, before feed forward networks and MLPs (\emph{orange}) generate the output. CvT additionally uses convolutional embedding (\emph{pink}). Swin transformer blocks (\emph{gray}) employ shifted windowing. All models conclude with class label outputs (labels are parameters used during the generation of the image.)%
  }
\end{figure*}

Convolutional vision transformer (CvT)~\cite{wu2021cvt} is a realization of a vision transformer in which convolutions are added to the structure, such that robustness and performance can be achieved at a lower computational cost. This architecture was designed to have the benefits of CNNs, such as local receptive fields, shared weights, spatial subsampling, as well as the benefits from transformers, such as dynamic attention, context fusion, and better generalization. The CvT architecture~\cite{wu2021cvt} differs from vViTs~\cite{dosovitskiy2020image} in two ways as shown in Fig.~\ref{fig_cvt}. First, to have CNN-like benefits, convolutional token embedding is applied to get the feature maps of the images. This also allows having a hierarchical multi-stage structure of transformers.
Second, convolution projection is applied to every self-attention block instead of linear projection. This is done as a depth-wise convolution operation. Here query, key, and values can also be downsampled using stride within the convolution projection without any performance loss. To classify the image, the fully connected MLP head is added at the last stage. The trainable parameters for CvT amount to approximately 19.55 million.

\subsubsection{Swin Transformer}
The Swin transformer introduced by Liu et al. employs hierarchical processing, by dividing images into smaller patches across multiple scales, and utilizes shifted windows to enhance the receptive field of each patch without increasing computational complexity~\cite{liu2021swin}. By incorporating transformer blocks within each stage as illustrated in Fig.~\ref{fig_cvt}, it effectively models interactions between patches, and captures contextual relationships. Swin transformers have shown competitive performance on tasks like image classification and object detection while maintaining computational efficiency. Like vViTs and CvTs, Swin transformers utilize transformer blocks within each stage to model interactions between patches and capture contextual relationships. However, they adaptively adjust the receptive field of each patch using shifted windows, enabling more efficient utilization of computational resources compared to vViTs. Unlike vViTs and CvTs, Swin transformers excel in capturing both local and global information efficiently. With a balanced approach to scalability and performance, they offer a promising solution for various applications.
The trainable parameters for Swin transformer amount to approximately 27.48 million. 

\begin{table}[H]
\caption{Comparison of the three tested ViTs architectures: vViT, CvT, and Swin. 
The table highlights key architectural distinctions, including the approximate number of trainable parameters, patch embedding, and the type of attention mechanism employed. 
While vViT adopts a simple fixed patch embedding with global attention, CvT introduces convolutional embeddings to better preserve spatial locality. 
Swin further departs from the global attention paradigm by employing a hierarchical structure with windowed attention.}
\begin{tabular}{p{1.5cm}p{1.5cm}p{2cm}p{2cm}}
\hline
Model      & Parameters     & Patch Style   & Attention Mechanism \\ \hline
vViT &  $\sim$5.7M   & Fixed         & Global \\ 
CvT  &  $\sim$19.6M   & Convolutional & Global \\ 
Swin   & $\sim$27.5M   & Hierarchical  & Windowed \\ \hline
\end{tabular}
\label{tab:vit-swin-cvt}
\end{table}
\subsection{Training Procedure}
Rather than relying on pretrained models, we chose to train each ViT architecture from scratch on our task-specific dataset. This approach provides two key advantages: first, it ensures that the models are directly optimized for the low-level perceptual tasks under investigation, rather than inheriting biases from unrelated pretraining objectives; second, it offers greater control over model inputs, architecture, and training dynamics—essential for fair comparison and for interpreting model behavior in a perceptual context.

To ensure, that we were able to obtain state-of-the-art results from the trained ViT architectures, our training procedure follows widely accepted training standards. For network initialization, we have employed Xavier's scheme~\cite{glorot2010xavier}, to ensure stable convergence. A batch size of $32$ was chosen, as it strikes a balance between computational efficiency and gradient accuracy, enabling effective weight updates without excessive memory usage. A stochastic gradient descent optimizer is employed for training, with a momentum coefficient set to $0.9$. This setup helps by dampening oscillations and facilitating faster convergence towards minima. The learning rate was set to $0.0001$, in order to control the step size of parameter updates during optimization. To prevent overfitting, a weight decay of $1 \times 10^{-6}$ is applied. Weight decay introduces an additional regularization term to the loss function, penalizing large weights and encouraging the model to generalize better to unseen data.

All models are trained on our cluster which comprise a variety of NVIDIA GPUs such as 3090, 4090 and A100. The training duration varies across different architectures on the 3090 and 4090 GPUs: vViT networks typically converge within 1-2 days, CvT networks require around 2-3 days, and Swin transformers typically take 3-4 days to converge. However, on the A100 GPU, the average training duration varies slightly, with ViTs taking approximately 1-1.5 days to converge. All training data stimuli generation is accomplished using the Python scikit-image library~\cite{van2014scikit,van1995python}, while training itself is conducted using PyTorch~\cite{paszke2019pytorch}.

\noindent\textbf{Confidence Intervals. } 
To quantify the variability of model performance, we report 95\% confidence intervals under the assumption of approximate normality of errors. 
Specifically, intervals are calculated as the sample mean plus or minus 1.96 times the sample standard deviation, which corresponds to the standard normal coverage for a two-sided 95\% interval. 
This method provides a conventional and interpretable measure of dispersion that allows for straightforward comparison across models and tasks.

\section{Experiments and Results}
In this section, we detail the experiments conducted for our investigation into low-level visual tasks. Subsequently, we present the results of these experiments in terms of quantitative measures and comparative analysis. In particular, we evaluate perceptual error using the mean log absolute error (MLAE) and define \emph{perceptual bias} as the systematic deviation between predicted and ground-truth perceptual responses. This definition provides the basis for the quantitative comparisons reported below. Our analysis involves a comparative evaluation of ViTs' performance against established benchmarks, in comparison to human observers and CNNs, as documented in prior research~\cite{cleveland1984graphical, haehn2018evaluating}.

\subsection{Performance Analysis of Humans vs. ViTs}
\begin{figure}[H]
  \centering 
  \includegraphics[width=.95\columnwidth]{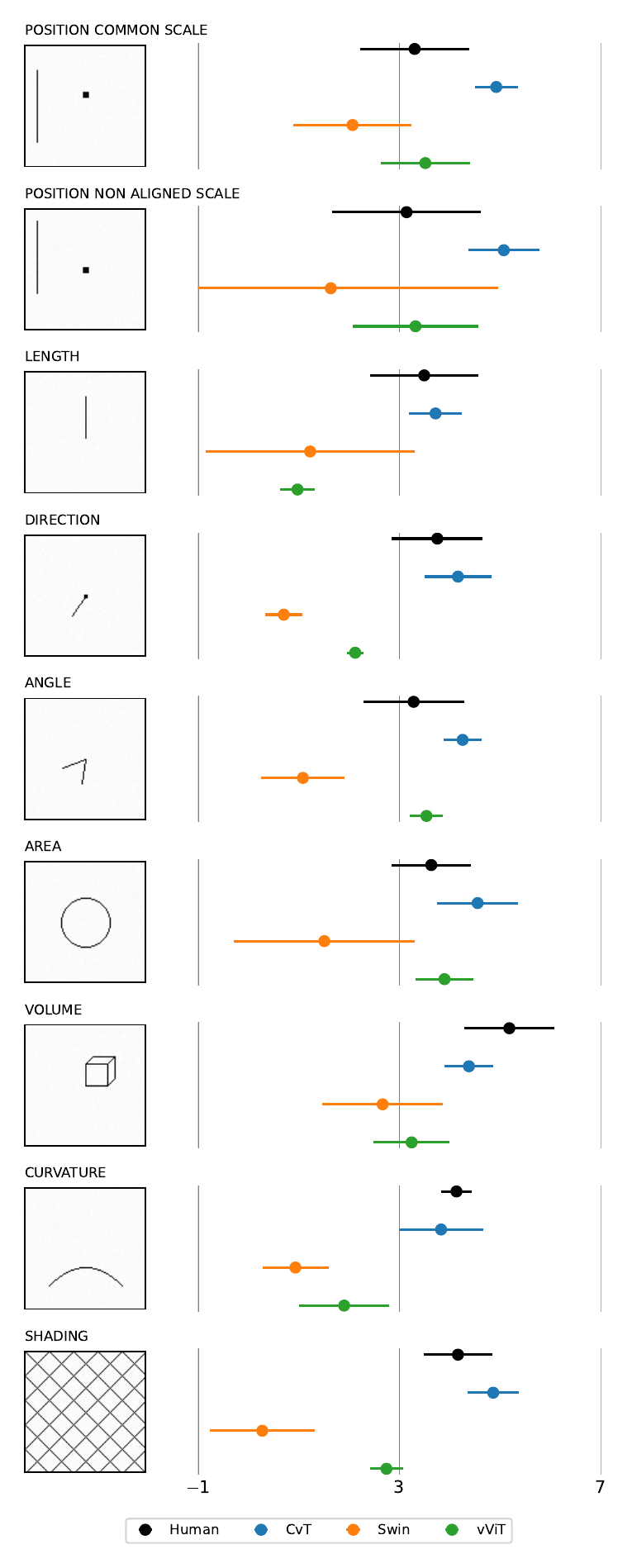}
  \caption{\label{figure_ele}%
  	Comparison of Regression Errors for Elementary Perceptual Tasks: elementary perceptual tasks experiment. Comparison of visual stimuli (left) and MLAE scores with  95$\%$ confidence intervals (lower is better) for various ViTs and humans (right). Swin generally aligns more closely with human performance for simpler encodings, while all ViTs diverge on complex features like curvature and shading. Human scores derived from Cleveland and McGill, and Hahen et al.'s studies~\cite{cleveland1984graphical, haehn2018evaluating}. %
  }
\end{figure}

\begin{figure}[H]
  \centering 
  \includegraphics[width=0.8\columnwidth]{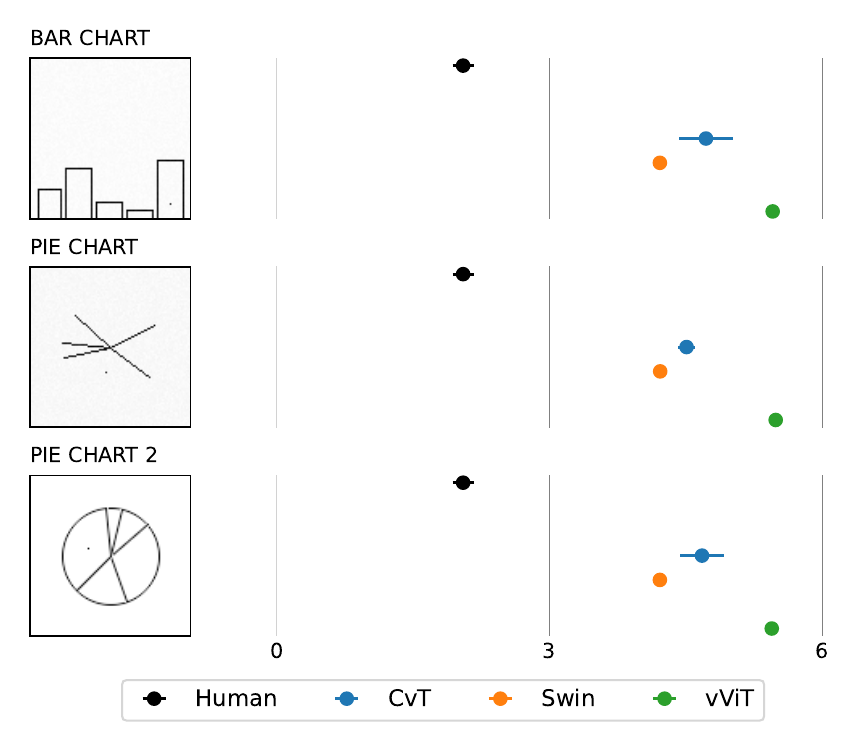}
  \caption{\label{figure_pa}%
  	Performance in Position-Angle Estimation: Comparison of visual stimuli (left) and MLAE scores with  95$\%$ confidence intervals (lower is better) for various ViTs and humans (right). Human scores derived from Cleveland and McGill, and Hahen et al.'s studies~\cite{cleveland1984graphical, haehn2018evaluating}.
    All ViTs underperform compared to human baselines.
  }
\end{figure}

\begin{figure}[H]
  \centering 
  \includegraphics[width=0.8\columnwidth]{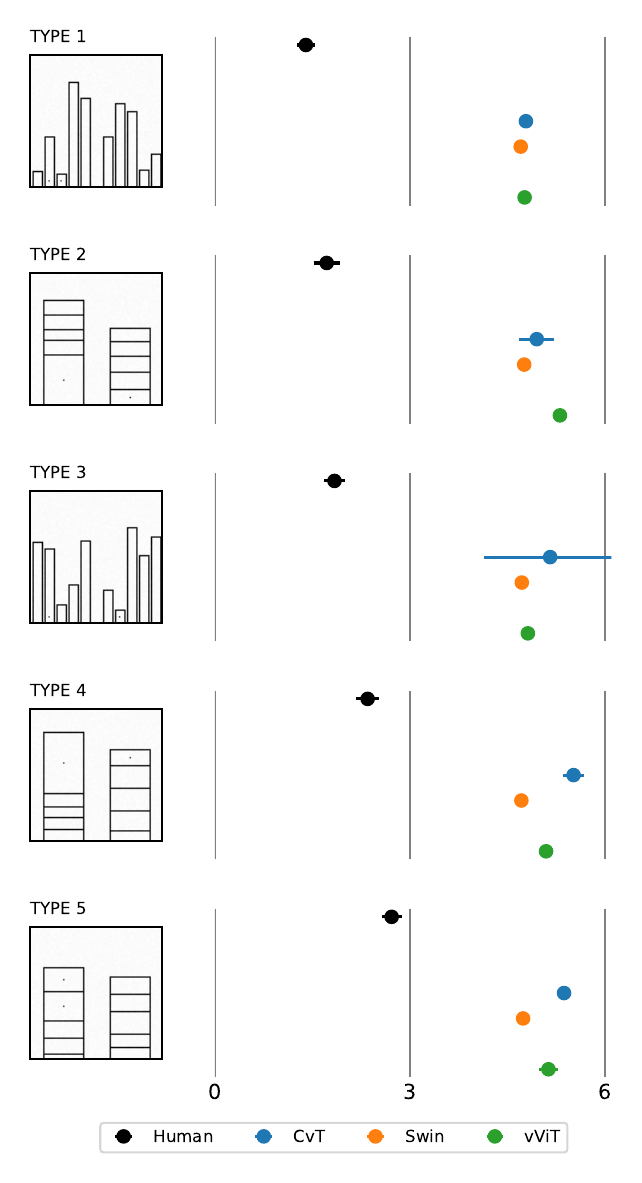}
  \caption{\label{figure_pl}%
  	Accuracy in Position-Length Estimation Across Chart Types: Comparison of visual stimuli (left) and MLAE scores with  95$\%$ confidence intervals (lower is better) for various ViTs and humans (right). Human scores derived from Cleveland and McGill, and Hahen et al.'s studies~\cite{cleveland1984graphical, haehn2018evaluating}. Swin outperforms other ViTs but still lags behind human perception. 
  }
\end{figure}

\begin{figure}[H]
  \centering 
  \includegraphics[width=0.80\columnwidth]{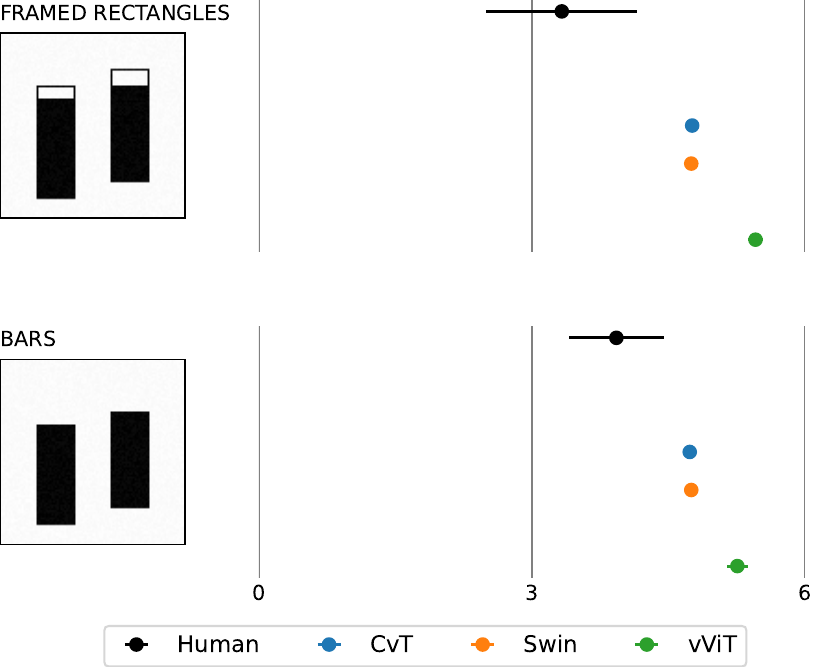}
  \caption{\label{figure_bf}%
  	Performance in Length Estimation with Framed Rectangles and Bars: Comparison of visual stimuli (left) and MLAE scores with  95$\%$ confidence intervals (lower is better, best result is marked as boldface) for various ViTs and humans (right). Human scores derived from Cleveland and McGill, and Hahen et al.'s studies~\cite{cleveland1984graphical, haehn2018evaluating}. Like in previous cases, humans perform better than ViTs.
  }
\end{figure}

\begin{figure}[H]
  \centering 
  \includegraphics[width=0.80\columnwidth]{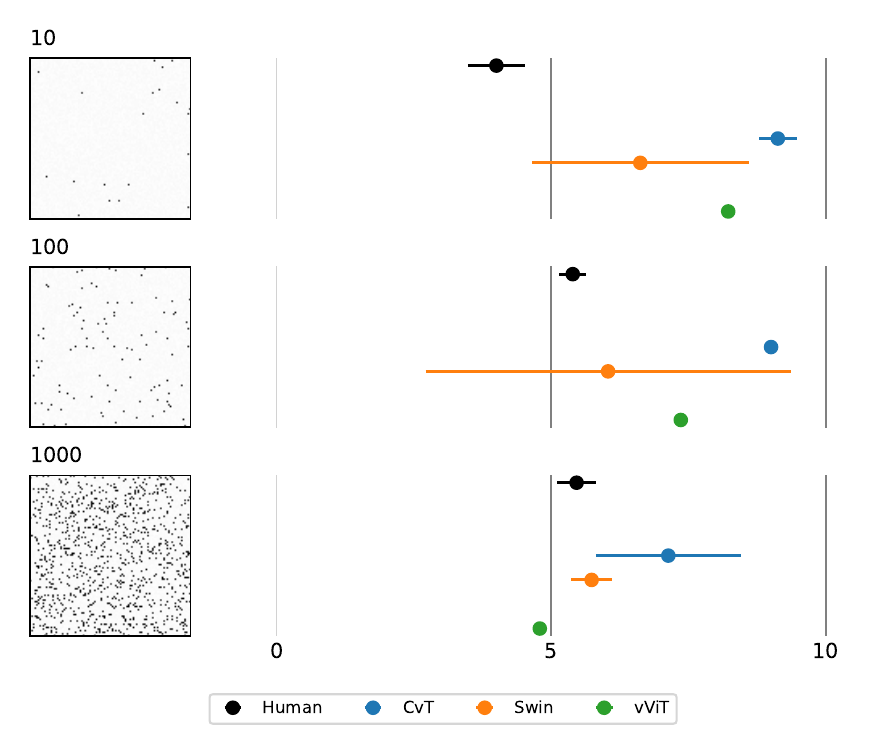}
  \caption{\label{figure_wb}%
  	 Perception of Quantity in Point Cloud Estimation. Comparison of visual stimuli (left) and MLAE scores with  95$\%$ confidence intervals (lower is better) for various ViTs and humans (right). Human scores derived from Cleveland and McGill, and Hahen et al.'s studies~\cite{cleveland1984graphical, haehn2018evaluating}.
     Note how ViTs struggle significantly with estimating dot quantities.
  }
\end{figure}

First, we conduct an experimental investigation into the performance of CvT, Swin, and vViT architectures across the nine basic encodings, utilizing mean squared error (MSE) for network regression. Following Cleveland and McGill~\cite{cleveland1984graphical} and Haehn et al.~\cite{haehn2018evaluating}, we use the midmean logistic absolute error (MLAE) to quantify perceptual accuracy. This metric is defined as:
\begin{equation}\label{eqn:eq_mlae}
    \text{MLAE} = log_2(|\text{predicted percent} - \text{true percent}| + 0.125)
\end{equation}
To evaluate perceptual accuracy, we compare MLAE results and analyze the average performance against humans. Fig.~\ref{figure_ele} depicts the results of regression analysis of the elementary perceptual tasks across different ViTs architectures and humans. Figures~\ref{figure_pa}, \ref{figure_pl}, \ref{figure_bf}, and \ref{figure_wb} show the results of our regression analysis of Position-Angle, Position-Length, Bars and Framed Rectangles and the Point Cloud task experiments respectively. The mean MLAE scores for ViTs  for each task are provided in Tables~\ref{tab:human_classifier_stats_ele}, ~\ref{tab:human_classifier_stats_pa}, ~\ref{tab:human_classifier_stats_pl}, ~\ref{tab:human_classifier_stats_bfr}, and~\ref{tab:human_classifier_stats_weber}. The mean average human scores are taken from Hahen et al.'s study~\cite{haehn2018evaluating}. 

As shown in Table~\ref{tab:table_combination}, human participants consistently outperformed top-performing ViT in most perceptual categories. For instance, in the Position-Length task category, humans achieved an average MLAE of 2.01, while Swin recorded a substantially higher error of 4.72. Similarly, in the Point Cloud estimation task, which tests sensitivity to density and distribution (subject to Weber’s Law), human observers achieved an MLAE of 4.95 compared to Swin’s 6.37. Notably, humans demonstrated particularly strong performance on the Bars and Framed Rectangles task (MLAE = 1.93), suggesting superior precision in comparative length estimation along non-aligned scales—something ViTs struggle with. These comparisons highlight that although ViTs excel in select categories like direction or shading, they still lack the holistic perceptual consistency exhibited by humans.

\begin{figure}[H]
  \centering \includegraphics[width=\columnwidth]{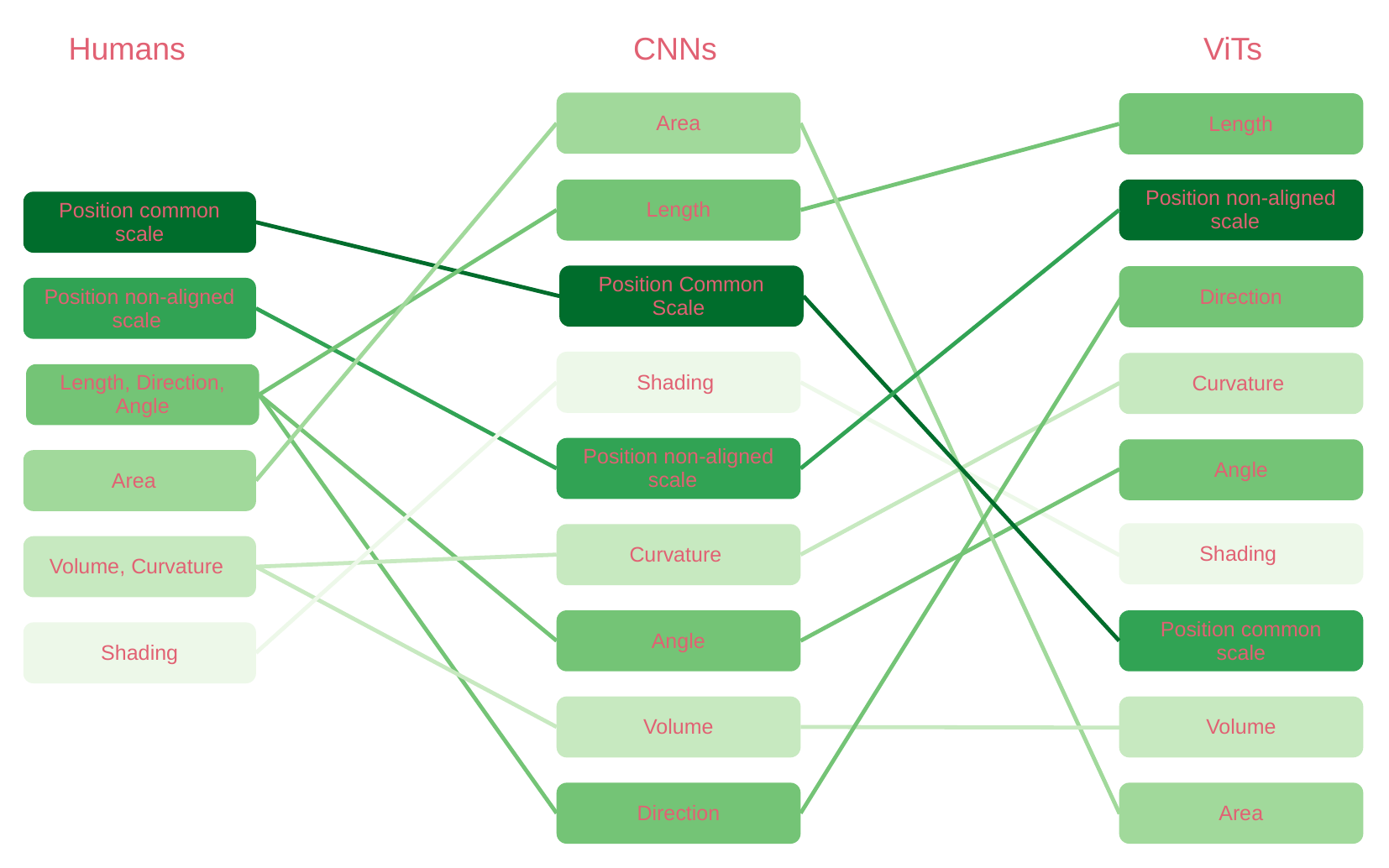}
  \caption{\label{fig:figure_ranking}% 
    Comparison of elementary perceptual task rankings across humans, CNNs, and ViTs. Rankings are derived based MLAE, with lower positions indicating better perceptual accuracy. The shading intensity of each box reflects rank: darker tones indicate higher accuracy (that is, rank 1), while lighter tones correspond to lower ranks. Human rankings are based on Cleveland and McGill~\cite{cleveland1984graphical}, while CNNs randking from Haehn et al.~\cite{haehn2018evaluating}, and ViT rankings are based on our retrained models.  Lines connect identical perceptual tasks across the three groups. Notably, ViTs exhibit most deviation—particularly in their treatment of position and length—highlighting greater ambiguity and weaker perceptual alignment with human judgments.
    }
\end{figure}

To better understand how ViTs preserve perceptual hierarchies, we compare task difficulty rankings based on MLAE. Elementary perceptual task ranking is a foundational aspect that involves ordering their accuracy relative to one another, where a task is considered more precise if it exhibits judgments that closely correspond with the actual encoded quantities. Fig.~\ref{fig:figure_ranking} contrasts the top ViT's perceptual task ordering with that of CNNs and humans. We present a comparison between the performance ranking of the top-performing network (that is, the one with the lowest MLAE) and the human baseline rankings taken from Cleveland and McGill~\cite{cleveland1984graphical} in Fig.~\ref{fig:figure_ranking}. Though all models show general agreement on the easiest tasks (such as length and position), ViTs diverge on more ambiguous encodings such as curvature and area.
%%%%%%%%%%% ranking figure %%%%%%%%%%%%%%%%%%
%%%%%%%%%%% ranking figure %%%%%%%%%%%%%%%%%%

\subsection{Performance Analysis of CNNs vs. ViTs}
To evaluate perceptual accuracy in low-level visual tasks, we conduct a comparative analysis of MLAE results and present the average performance across CNNs and ViTs.
The CNN architectures used in our analysis follow those evaluated by Haehn et al.~\cite{haehn2018evaluating}, namely LeNet, VGG19, and Xception, with the addition of ResNet-18~\cite{he2016deep}. The regression error comparison for CNNs and ViTs for elementary perceptual tasks is reported in Table~\ref{tab:trained_models_ele}. 
Across all tasks, ViTs consistently exhibited higher error rates than CNNs, with the largest performance gap observed in the point cloud estimation task, where the average MLAE for ViTs was 6.37 compared to 3.40 for CNNs. Similarly, for bar and framed rectangle tasks, ViTs yielded significantly higher errors (4.75 vs. 1.93), suggesting weaker performance in estimating lengths across non-aligned scales. The smallest gap was observed in position-angle tasks, with Swin achieving an average error of 4.21 compared to 2.93 for VGG19. These results highlight that while ViTs can approximate relative encodings, their precision in absolute perceptual judgments still lags behind more specialized CNNs.
Table~\ref{tab:table_combination} illustrates the evaluation of CNNs and ViTs across our various other tasks (see Section~\ref{low}). 
Fig.~\ref{fig:figure_ranking} visualizes the task ranking based on MLAE for humans~\cite{cleveland1984graphical}, the best-performing CNN (VGG19) from Haehn et al.~\cite{haehn2018evaluating}, and our best ViT model (Swin). ViTs rank these encodings as more difficult than both humans and CNNs.

Additionally, we compare the average regression performance of the ViT architectures across all evaluated tasks.
This aggregated analysis provides a clearer view of how the models perform relative to one another beyond single-task outcomes.
To ensure the robustness of these findings, we conduct statistical tests and report effects that are significant.

\noindent\textbf{Elementary perceptual task.} Across elementary perceptual tasks, among all networks, there is a statistically significant difference ($F=14.20$, $p<0.001$). Tukey HSD tests~\cite{abdi2010tukey} indicate that Swin performed significantly better than vViT ($t=1.19$, $p<0.001$) and CvT performed significantly better than vViT ($t=0.81$, $p=0.0001$), while Swin and CvT did not differ significantly.

\noindent\textbf{Position-angle task.} ANOVA~\cite{st1989analysis} showed a strong main effect ($F=419.14$, $p<0.001$). Tukey HSD tests revealed that Swin performed significantly better than both CvT ($t=0.42$, $p<0.001$) and vViT ($t=1.25$, $p<0.001$). In addition, CvT was significantly better than vViT ($t= 0.83$, $p<0.001$).

\noindent\textbf{Position-length task.} ANOVA revealed a significant effect among the three networks ($F=11.44$, $p<0.001$). Tukey HSD tests indicated that Swin performed significantly better than CvT ($t=0.43$, $p<0.001$) and vViT ($t=0.29$, $p=0.007$), while CvT and vViT did not differ significantly.

\noindent\textbf{Bar and framed rectangle task.} ANOVA also indicated a significant effect ($F=133.55$, $p<0.001$). Tukey HSD comparisons showed that both Swin ($t=0.61$, $p<0.001$) and CvT ($t=0.61$, $p<0.001$) performed significantly better than vViT, whereas CvT and Swin did not differ significantly.

\noindent\textbf{Point cloud task.} ANOVA yielded a significant effect ($F=7.31$, $p<0.0033$). Tukey HSD results demonstrated that CvT performed significantly worse than both Swin ($t= -2.29$, $p<0.003$) and vViT ($t=-1.63$, $p<0.037$), while Swin and vViT did not differ significantly.

%%%%%%%%%% table for pl, pa, bfr, weber %%%%%%%%%%
\begin{table}[H]
  \caption{%
     Average error (MLAE) for humans, and the best-performing CNN (VGG19) from Haehn et al.~\cite{haehn2018evaluating}, and our best ViT model (Swin) different perceptual experiments. All models were trained from scratch under a unified setup for fair comparison. Lower values indicate better perceptual accuracy.
  }
  \label{tab:table_combination}
  \centering
  \begin{tabular}{p{3cm} p{1.5cm} p{1.5cm} p{1cm}}
    \toprule
    Task Category & Humans & VGG19~\cite{haehn2018evaluating} & Swin \\
    \midrule
    \multicolumn{4}{l}{\textit{Position-Angle}} \\
    Bar & & 2.18 & 4.22 \\
    Pie without outline & & 3.30 & 4.21 \\
    Pie & & 3.30 & 4.21 \\
    \midrule
    Average Error & \textbf{2.05} & 2.93 & 4.21 \\
    \midrule
    \multicolumn{4}{l}{\textit{Position-Length}} \\
    Type 1 & & 3.96 & 4.72 \\
    Type 2 & & 3.95 & 4.74 \\
    Type 3 & & 4.35 & 4.72 \\
    Type 4 & & 3.67 & 4.73 \\
    Type 5 & & 3.90 & 4.71 \\
    \midrule
    Average Error & \textbf{2.01} & 3.97 & 4.72 \\
    \midrule
    \multicolumn{4}{l}{\textit{Bars and Framed Rectangles}} \\
    Bars & & 1.98 & 4.76 \\
    Framed Rectangles & & 1.87 & 4.74 \\
    \midrule
    Average Error & 3.63 & \textbf{1.93} & \textbf{4.75} \\
    \midrule
    \multicolumn{4}{l}{\textit{Point Cloud}} \\
    Base 10 & & 1.65 & 5.21 \\
    Base 100 & & 3.84 & 8.43 \\
    Base 1000 & & 4.71 & 5.48 \\
    \midrule
    Average Error & 4.95 & \textbf{3.40} & 6.37 \\
    \bottomrule
  \end{tabular}
\end{table}
%%%%%%%%%% table for pl, pa, bfr, weber %%%%%%%%%%%%%
%%%%%%%%%%%%%Figure for cross_network %%%%%%%%%%%
\begin{figure}[H]
  \centering 
  \includegraphics[width=.9\columnwidth]{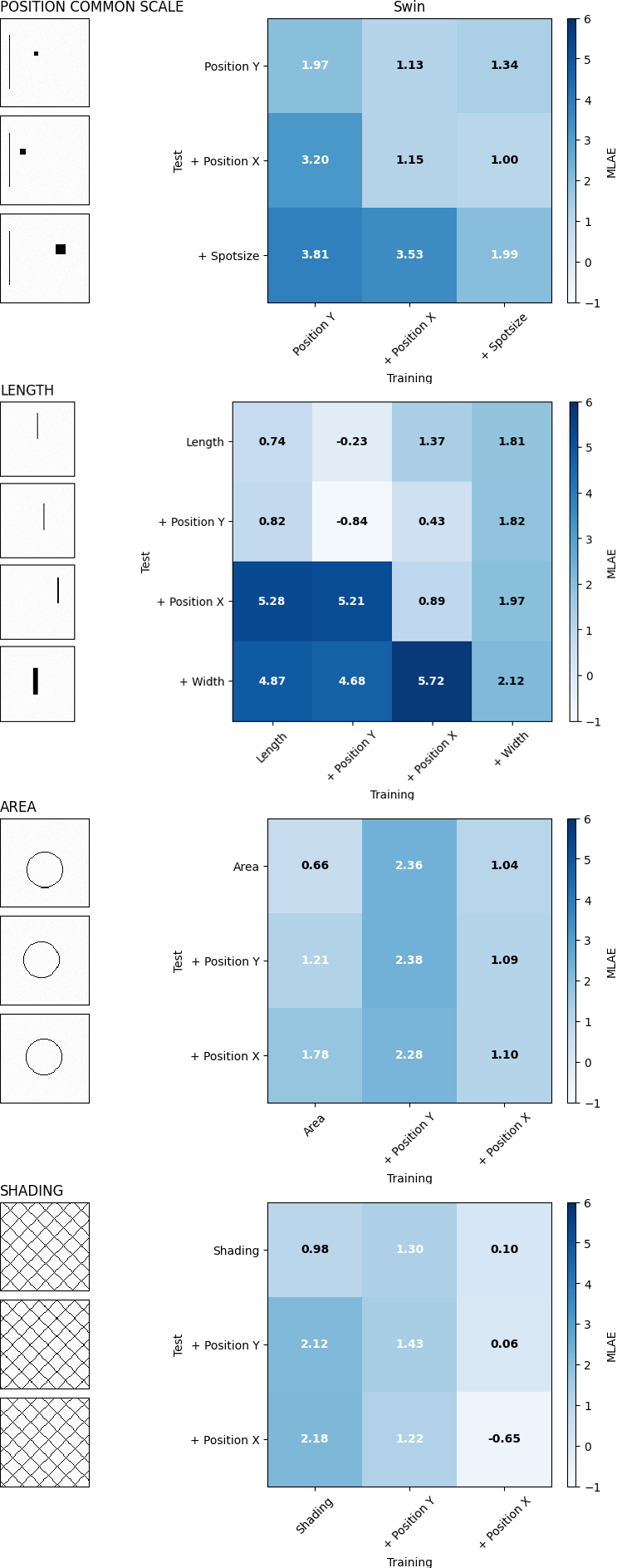}
  \caption{\label{figure_cn_swin}%
  Cross-generalization performance of the Swin Transformer across different parameterizations of elementary perceptual tasks. Each matrix corresponds to a specific task (for example, Position, Length, Area, Shading). Rows represent the parameterization used for training, while columns represent the parameterization used for testing. The top row in each matrix shows within-task (in-parameter) performance, with diagonal entries indicating baseline accuracy when training and testing on the same parameter configuration. Off-diagonal entries indicate the model’s ability to generalize to new parameter settings (that is, changes in object position, size, or shape). Higher values off-diagonal reflect poor generalization and sensitivity to unseen visual variations.
  }
\end{figure}
%%%%%%%%%%%%%%%%%%%%%%%%%%%
%%%%%%%%%%%%%%Figures with  errorbars %%%%%%%%%%%%
%%%%%%%%%%%%%%%%%%%%%%%%%%%%%%%%%%%%%%%%%%

%%%%%%%%%%%%%%Figures with  errorbars %%%%%%%%%%%% 

We present a selection of cross-parameterization results in Fig.~\ref{figure_cn_swin}. These matrices are shown only for our best-performing network, the Swin Transformer, to highlight upper-bound performance within the ViT family. Each matrix corresponds to a specific elementary task (for example, position, length, area, shading), with rows denoting the parameterization used during training and columns representing those used during testing. Diagonal entries capture in-task performance, while off-diagonal entries reveal generalization to unseen variations. The consistently elevated MLAE values off-diagonal suggest that Swin struggles to generalize across modest changes in visual configuration—such as variations in object width, spatial alignment, or density. This pattern is especially pronounced in the \textit{Position} and \textit{Length} tasks, where generalization breaks down significantly. These results indicate that even top-performing ViTs are highly sensitive to training conditions, limiting their applicability in settings requiring robust perceptual inference.

\begin{figure}[H]
  \centering 
  \includegraphics[width=0.9\columnwidth]{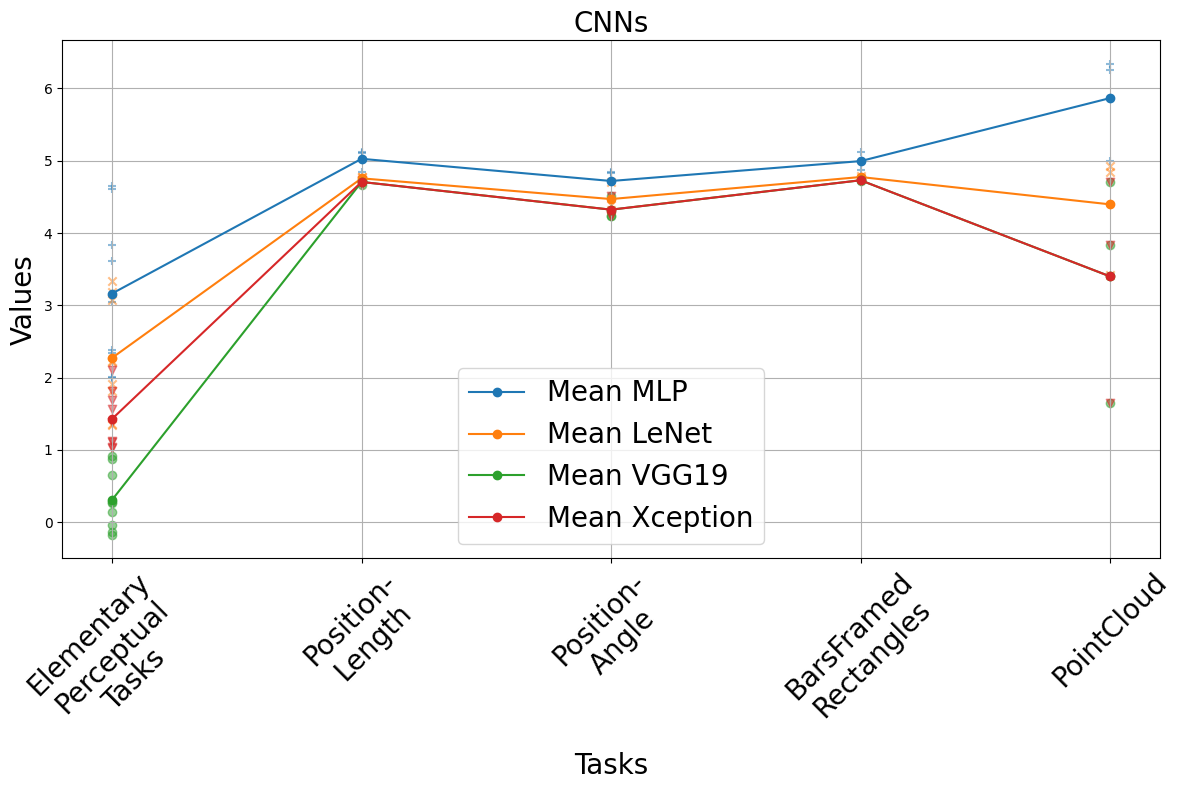}
  \caption{%
  	Analysis of mean performance for CNNs from the study by Hahen et al.~\cite{haehn2018evaluating} across five experiments. The lowest mean score indicates the best-performing CNN.%
  }
  \label{fig:figure_cnn_mean}
\end{figure}

\begin{figure}[H]
  \centering 
  \includegraphics[width=0.9\columnwidth]{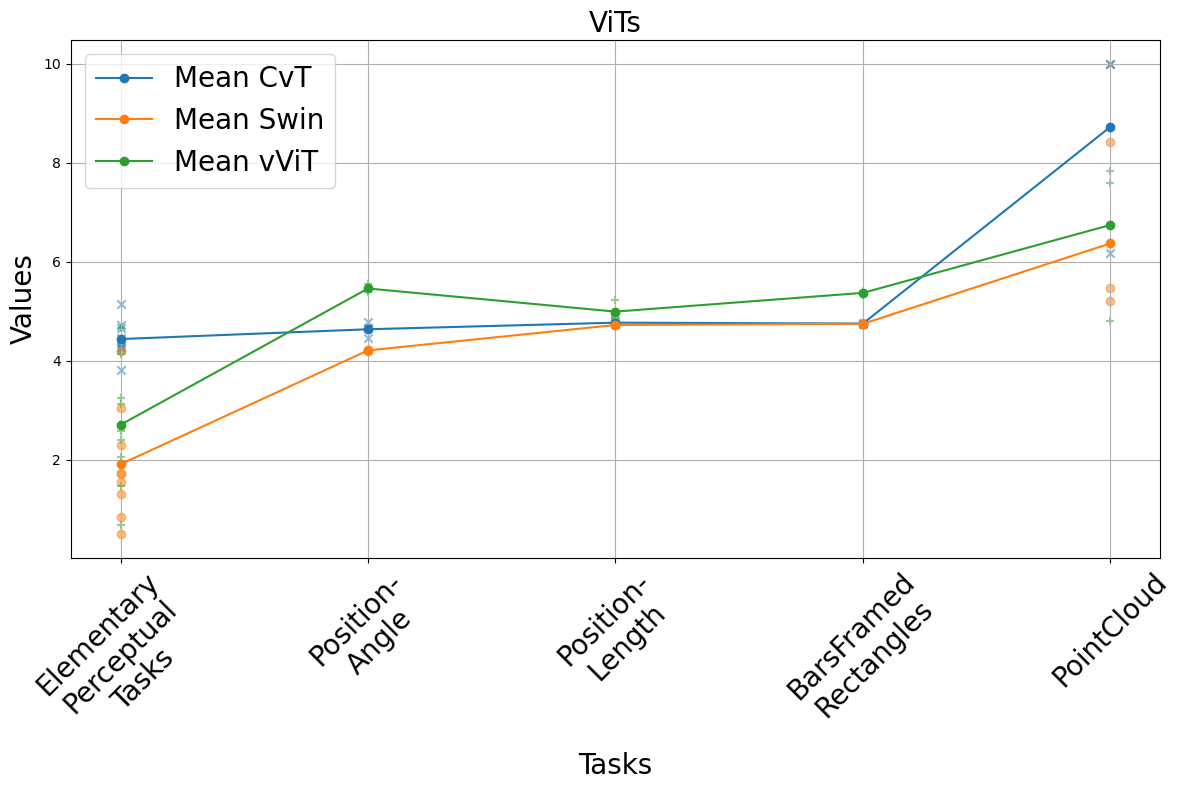}
  \caption{%
   Analysis of mean performance for our trained ViTs across five experiments. Swin shows the best overall performance among ViTs, yet remains inconsistent compared to CNNs in several key encodings.%
  }
  \label{fig:figure_transformers_mean}
\end{figure}

Figures~\ref{fig:figure_cnn_mean} and~\ref{fig:figure_transformers_mean} illustrate the comparison of mean performance between CNNs and ViTs for each task. They showcase the best-performing CNN and ViT individually in separate plots. While Swin leads among ViTs, it remains inconsistent and trails behind CNNs in key perceptual benchmarks.

%%%%%%%%%%%%%%%  Figures with average plots %%%%%%%%%%%%%%%%%

%%%%%%%%%%%%%%%  Table for cnn and vit %%%%%%%%%%%%%%%%%
%% table for different cnns and transformers for low level tasks 
\begin{table*}
  \caption{%
  	Performance comparison on elementary perceptual tasks for various CNN and ViT models. The comparison is based on the MLAE scores (lower is better, best results encoded in boldface). VGG19 consistently achieves the lowest error among CNNs, while Swin leads the ViTs, yet with noticeably higher errors in complex visual encodings.
  }
  \label{tab:trained_models_ele}
  % \scriptsize%
  \centering%
  \begin{tabular}{cccccccccccc}
  	\toprule
  	{Network} & \rotatebox{90}{Position} \rotatebox{90}{common } \rotatebox{90}{scale} & \rotatebox{90}{Position}  \rotatebox{90}{non-aligned}\rotatebox{90}{scale} & \rotatebox{90}{Length} & \rotatebox{90}{Direction} & \rotatebox{90}{Angle} & \rotatebox{90}{Area} &\rotatebox{90}{Volume} & \rotatebox{90}{Curvature}  & \rotatebox{90}{Shading} & & {Average Error}  \\
  	\midrule
  	MLP \cite{haehn2018evaluating} & 3.84 &  3.61 & 1.99 & 4.65 & 4.61 & 2.01 & 2.38 & 2.34 & 3.04 & & 3.16\\
    \midrule
  	LeNet \cite{haehn2018evaluating} & 1.36 & 1.35 & 3.19 & 3.07 & 3.33 & 2.21 & 1.91 & 1.81 & 2.23 & &2.27\\
  	VGG19 \cite{haehn2018evaluating} & -0.04 & 0.26 & -0.14 & 0.92 & 0.66 & -0.17   & 0.87 & 0.28 & 0.14 & & \textbf{0.30}\\
  	Xception \cite{haehn2018evaluating}  & 1.04 & 1.02  & 1.11 & 1.57 & 1.69   & 1.38 & 2.10   & 1.13 & 1.82 & & 1.42 \\
        ResNet-18& 2.33  & 3.22  & 1.65 & 2.21 & 3.71   & 3.01 & 2.46   & 1.71 & 2.73 & &2.55 \\
        \midrule
          &   &   &  &  &     &   &   & & & Average CNNs Error &\textbf{1.64}\\
        \midrule
  	CvT & 4.76 & 5.40 & 3.28 & 3.62 & 4.15 & 4.65 & 4.60 & 4.19 & 5.19 & & 4.42 \\
  	Swin & 1.85 & -0.56  & -1.38 & 0.57 & 0.83  & 2.94   & 2.54  & 0.82 & 0.95 & & \textbf{0.95}\\
       % Swin (S) & 1.40 & 1.54  & 1.66 & 0.92 & 4.29  & 2.27  & 3.99   & 2.27 & 0.068 && 2.04 \\
  	vViT & 3.25& 3.03& 1.26& 2.07& 3.48& 4.06& 2.81& 2.51& 2.46&&  2.77\\
        %DeIT &3.30 &4.38&1.13&1.51&2.80&3.92&3.54&2.50&3.44 & &2.94\\
  	\midrule   
          &   &   &  &  &     &   &   & & & Average ViTs Error &2.71\\
        \midrule
  \end{tabular}%
\end{table*}
%%%%%%%%%%%%%%%  Table for cnn and vit %%%%%%%%%%%%%%%%%

%%%%%%%%%%%  Table for weighted cnn and vit %%%%%%%%%%%%%%%%
\begin{table*}
  \caption{%
  	Performance comparison of CNNs and ViTs (* indicates pretrained on ImageNet~\cite{NIPS2012_c399862d}) on elementary perceptual tasks (lower is better, boldface indicates the best performing model). Pretrained CvT shows improved perceptual accuracy compared to models trained from scratch.
  }
  \label{tab:pre_trained_models_ele}
  % \scriptsize%
  \centering%
  \begin{tabular}{cccccccccccc}
  	\toprule
  	{Network} & \rotatebox{90}{Position} \rotatebox{90}{common } \rotatebox{90}{scale} & \rotatebox{90}{Position}  \rotatebox{90}{non-aligned}\rotatebox{90}{scale} & \rotatebox{90}{Length} & \rotatebox{90}{Direction} & \rotatebox{90}{Angle} & \rotatebox{90}{Area} &\rotatebox{90}{Volume} & \rotatebox{90}{Curvature}  & \rotatebox{90}{Shading} &  & {Average Error}  \\
  	\midrule
        VGG19{*} \cite{haehn2018evaluating} & 1.02 & 1.09 & 0.87 & 2.84 & 2.31 & 0.49  & 1.16 & 0.71 & 0.73 & & \textbf{1.24}\\
        Xception{*} \cite{haehn2018evaluating} & 1.65 & 1.71 & 1.59 & 3.46 & 2.60 & 0.80  & 2.03 & 1.17 & 1.57 & & 1.84\\
        \midrule
          &   &   &  &  &     &   &   & & & Average CNNs Error &\textbf{1.54}\\
        \midrule
        CvT{*}  & 2.55 & 3.64 & -0.38 & 1.18 & 0.88 & 0.62  & 0.62 & 4.03 & 2.38 && \textbf{1.70}\\
        Swin{*}  & 2.10 & 2.16 & 1.93 & 1.38 & 0.55 & 1.53  & 3.54 & 1.94 & 3.55 & & 2.15\\
        vViT{*}  & 0.97 & 3.28 & 0.03 & 1.99 & 2.39 & 1.33  & 4.25 & 0.90 & 2.98 & & 2.01\\
        %DeIT{*} & 0.94& 3.37&0.14&2.01&2.27&1.42&4.14&0.94&2.98&&2.02\\
        \midrule   
          &   &   &  &  &     &   &   & & & Average ViTs Error &1.95\\
  	\midrule               
  \end{tabular}%
\end{table*}
%%%%%%%%%%%%%%% Table for ablation on vit %%%%%%%%%%%%%%%%%
\begin{table*}
  \caption{%
    Average MLAE scores (lower is better) of three ViT models (CvT, Swin, vViT) across various experimental conditions in our ablation study: base architecture, increased image resolution, expanded training data, and smaller patch sizes (vViT-8). Results show minimal performance gains from increased data or resolution. Swin consistently outperforms others, especially in elementary perceptual tasks, while vViT-8 shows improvement only in specific configurations like position-length.%
  }
  \label{tab:ablation_study}
  \centering
  \setlength{\tabcolsep}{4pt}
  \begin{tabular}{lcccccccccccccc}
    \toprule
    Experiments & \multicolumn{3}{c}{{Base}} & \multicolumn{3}{c}{{Image Resolution}} & \multicolumn{3}{c}{{Large Data}} & \multicolumn{3}{c}{{Pretraining}} & {Small Patch} \\
    \cmidrule(lr){2-4} \cmidrule(lr){5-7} \cmidrule(lr){8-10} \cmidrule(lr){11-13} \cmidrule{14-14}
     & CvT & Swin & vViT & CvT & Swin & vViT & CvT & Swin & vViT & CvT & Swin & vViT & vViT-8 \\
    \midrule
    Elementary Perceptual Tasks     & 4.42 & \textbf{0.95} & 2.77 & 4.44 & 2.60 & 3.05 & 4.02 & 1.99 & 2.58 & \textbf{1.70} & 2.15 & 2.01 & 4.61 \\
    
    Position-Length                 & 5.34 & 4.72 & 5.05 & 5.15 & 4.73 & 5.09 & 5.05 & 4.72 & 5.03 & 4.77 & 4.70 & 5.00 & \textbf{2.14} \\
    Positron-Angle                 & 4.62 & \textbf{4.21} & 5.41 & 4.93 & 4.22 & 5.54 & 4.66 & 4.21 & 5.30 & 4.47 & 4.32 & 4.72 & 5.51 \\
    Bars and Framed Rectangles     & \textbf{4.72} & 4.75 & 5.31 & 4.80 & 4.77 & 5.42 & 4.78 & 4.76 & 5.13 & 4.78 & 4.73 & 4.99 & 5.35 \\
    Point Cloud                    & 9.09 & 6.02 & 6.83 & 35.21 & 4.80 & 7.01 & 42.41 & 6.77 & 6.60 & 4.39 & \textbf{3.94} & 5.86 & 7.45 \\
    \bottomrule
  \end{tabular}
\end{table*}

%%%%%%%%%%%%%% table for meand and std fro elementary perceptual tasks %%%%%%%%%%%%%%%%
\begin{table*}
\caption{Comparison of mean MLAE and standard deviation between human observers and ViT models across nine elementary perceptual tasks. While ViTs like Swin achieve low average errors in certain tasks (e.g., shading, direction), they exhibit higher variance and inconsistency across encodings compared to humans, highlighting their limited reliability in generalizing perceptual judgments.}
\label{tab:human_classifier_stats_ele}
\centering
\begin{tabular}{lcccc}
\toprule
Category & Human (mean $\pm$ std)~\cite{haehn2018evaluating} & CvT (mean $\pm$ std) & Swin (mean $\pm$ std) & vViT (mean $\pm$ std) \\
\midrule
Position common scale      & 3.30 $\pm$ 1.08 & 4.68 $\pm$ 0.06 & \textbf{2.81 $\pm$ 1.02} & 3.77 $\pm$ 0.53 \\
Position non-aligned scale & 3.14 $\pm$ 1.48 & 4.88 $\pm$ 0.26 & \textbf{2.61 $\pm$ 2.46} & 3.57 $\pm$ 0.47 \\
Length                     & 3.49 $\pm$ 1.08 & 3.77 $\pm$ 0.25 & 1.84 $\pm$ 1.04 & \textbf{1.23 $\pm$ 0.42} \\

Direction                  & 3.75 $\pm$ 0.90 & 4.14 $\pm$ 0.16 & \textbf{0.72 $\pm$ 0.24} & 2.12 $\pm$ 0.13 \\
Angle                      & 3.28 $\pm$ 1.00 & 4.16 $\pm$ 0.14 & \textbf{0.88 $\pm$ 0.61} & 3.48 $\pm$ 0.28 \\

Area                       & 3.63 $\pm$ 0.79 & 4.80 $\pm$ 0.51 & \textbf{2.25 $\pm$ 1.08} & 3.88 $\pm$ 0.11 \\
Volume                     & 5.18 $\pm$ 0.90 & 4.39 $\pm$ 0.14 & \textbf{2.67 $\pm$ 0.50} & 3.24 $\pm$ 0.38 \\
Curvature                  & 4.13 $\pm$ 0.30 & 4.18 $\pm$ 0.15 & \textbf{0.93 $\pm$ 0.49} & 2.15 $\pm$ 0.34 \\
Shading                    & 4.16 $\pm$ 0.68 & 4.87 $\pm$ 0.19 & \textbf{0.36 $\pm$ 0.74} & 2.74 $\pm$ 0.22 \\
\midrule
\end{tabular}
\end{table*}

%%%%%%%%%%%%%% table for meand and std position angle task %%%%%%%%%%%%%%%%
\begin{table*}
\caption{Comparison of mean MLAE and standard deviation between human observers and ViT models across position-angle tasks. While ViTs like Swin achieve low average errors in each position angle task but they exhibit higher errors compared to humans, highlighting their limited reliability in generalizing perceptual judgments.}
\label{tab:human_classifier_stats_pa}
\centering
\begin{tabular}{lcccc}
\toprule
Task & Human (mean $\pm$ std)~\cite{haehn2018evaluating} & CvT (mean $\pm$ std) & Swin (mean $\pm$ std) & vViT (mean $\pm$ std) \\
\midrule
Bar chart     & \textbf{2.05 $\pm$ 0.12} & 4.72 $\pm$ 0.19 & 4.21 $\pm$ 0.01 & 5.45 $\pm$ 0.03 \\
Pie chart     & \textbf{2.05 $\pm$ 0.12} & 4.51 $\pm$ 0.06 & 4.22 $\pm$ 0.01 & 5.49 $\pm$ 0.03 \\
Pie chart without outline & \textbf{2.05 $\pm$ 0.12} & 4.68 $\pm$ 0.15 & 4.21 $\pm$ 0.01 & 5.44 $\pm$ 0.05 \\
\midrule
\end{tabular}
\end{table*}
%%%%%%%%%%%%%% table for meand and std position length task %%%%%%%%%%%%%%%%
\begin{table*}
\caption{Comparison of mean MLAE and standard deviation between human observers and ViT models across position-length tasks. While ViTs like Swin achieve low average errors in each position length type and they exhibit higher errors compared to humans, highlighting their limited reliability in generalizing perceptual judgments.}
\label{tab:human_classifier_stats_pl}
\centering
\begin{tabular}{lcccc}
\toprule
Category & Human (mean $\pm$ std)~\cite{haehn2018evaluating} & CvT (mean $\pm$ std) & Swin (mean $\pm$ std) & vViT (mean $\pm$ std) \\
\midrule
Position length type 1 & \textbf{1.40 $\pm$ 0.14} & 4.79 $\pm$ 0.05 & 4.71 $\pm$ 0.02 & 4.77 $\pm$ 0.04 \\
Position length type 2 & \textbf{1.72 $\pm$ 0.20} & 4.95 $\pm$ 0.17 & 4.76 $\pm$ 0.04 & 5.31 $\pm$ 0.06 \\
Position length type 3 & \textbf{1.84 $\pm$ 0.16} & 5.16 $\pm$ 0.64 & 4.72 $\pm$ 0.01 & 4.82 $\pm$ 0.07 \\
Position length type 4 & \textbf{2.35 $\pm$ 0.18} & 5.52 $\pm$ 0.10 & 4.72 $\pm$ 0.01 & 5.10 $\pm$ 0.05 \\
Position length type 5 & \textbf{2.72 $\pm$ 0.16} & 5.37 $\pm$ 0.01 & 4.74 $\pm$ 0.01 & 5.13 $\pm$ 0.09 \\
\midrule
\end{tabular}
\end{table*}
%%%%%%%%%%%%%% table for meand and std bar and framed rectangles task %%%%%%%%%%%%%%%%
\begin{table*}
\caption{Comparison of mean MLAE and standard deviation between human observers and ViT models for bar and framed rectangle tasks. Swin achieve low average error for framed rectangle tasks and CvT for Bar tasks but they exhibit higher errors compared to humans, highlighting their limited reliability in generalizing perceptual judgments.}
\label{tab:human_classifier_stats_bfr}
\centering
\begin{tabular}{lcccc}
\toprule
Task & Human (mean $\pm$ std) & CvT (mean $\pm$ std) & Swin (mean $\pm$ std) & vViT (mean $\pm$ std) \\
\midrule
Framed Rectangle & \textbf{3.33 $\pm$ 0.83} & 4.76 $\pm$ 0.03 & 4.75 $\pm$ 0.02 & 5.46 $\pm$ 0.05 \\
Bar       & \textbf{3.93 $\pm$ 0.52} & 4.74 $\pm$ 0.05 & 4.75 $\pm$ 0.01 & 5.26 $\pm$ 0.07 \\
\midrule
\end{tabular}
\end{table*}
%%%%%%%%%%%%%% table for meand and std point cloud estimation task %%%%%%%%%%%%%%%%
\begin{table*}
\caption{Comparison of mean MLAE and standard deviation between human observers and ViT models across three point cloud estimation tasks. vViT achieve low average error in Base 1000 task but overall exhibit higher errors compared to humans, highlighting their limited reliability in generalizing perceptual judgments.}
\label{tab:human_classifier_stats_weber}
\centering
\begin{tabular}{lcccc}
\toprule
Task & Human (mean $\pm$ std)~\cite{haehn2018evaluating} & CvT (mean $\pm$ std) & Swin (mean $\pm$ std) & vViT (mean $\pm$ std) \\
\midrule
Base 10 & \textbf{4.00 $\pm$ 0.52} & 9.12 $\pm$ 0.21 & 6.62 $\pm$ 1.23 & 8.22 $\pm$ 0.07 \\
Base 100 & \textbf{5.39 $\pm$ 0.25} & 9.00 $\pm$ 0.00 & 6.03 $\pm$ 2.08 & 7.36 $\pm$ 0.08 \\
Base 1000 & 5.46 $\pm$ 0.35 & 7.13 $\pm$ 0.82 & 5.73 $\pm$ 0.23 & \textbf{4.79 $\pm$ 0.00} \\
\midrule
\end{tabular}
\end{table*}

%%%%%%%%%%%%%%%%%%%%%%%%%%% classifier's stats %%%%%%%%%%%%%%%%%
\section{Discussion}
To investigate the alignment of ViTs with humans' visual processing of visual data, we discuss and relate the individual results for ViTs, humans and also CNNs.
Our discussion draws on quantitative differences in perceptual error (MLAE), as well as task-specific rankings, to evaluate the strengths and limitations of each approach.

\subsection{Humans vs. ViTs}
Our investigations indicate that across the nine elementary perceptual tasks, ViTs demonstrated competitive performance compared to human observers. ViTs achieved an MLAE of 2.71 across elementary tasks, compared to 3.16 for human observers, with Swin achieving the lowest error of 0.95 (see Table~\ref{tab:trained_models_ele}). In particular, Swin outperformed human participants in direction estimation (MLAE: 0.72 vs. 3.75) and shading (0.36 vs. 4.16), suggesting strong capabilities in processing fine-grained texture and orientation cues.
Ranking analysis (see Fig.~\ref{fig:figure_ranking}) further indicates partial alignment between ViTs and human judgments. Best performing ViT model Swin exhibit similar alignment with human ranking for position non-aligned scale (rank:2), direction (rank:3) and shading (rank:6). Yet, the divergence became apparent in tasks involving curvature and area, where ViTs ranked these tasks as significantly easier than humans did, pointing to fundamental differences in how visual information is processed.
As previously reported, humans often face challenges with such tasks due to the lack of visual cues and spatial reasoning abilities~\cite{cleveland1984graphical}. In contrast, ViTs excel in solving these tasks more effectively, potentially by leveraging their inherent attention mechanisms. However, when comparing performance across tasks, such as position-length, position-angle, bars and framed rectangles, as well as point cloud perception, humans  outperform ViTs. 
For instance, in the position-length task(see Table~\ref{tab:table_combination}, human observers achieved an average MLAE of 2.01, whereas Swin’s error was more than double at 4.72. Similarly, in the point cloud task, Swin recorded an average MLAE of 6.37, compared to 4.95 for humans. This discrepancy prompts significant questions about the feasibility of using ViTs in the domain of visualization methodologies. 
%%%
These results underscore that while ViTs are effective in some perceptual judgments, particularly those involving local structure or texture, they remain less reliable in tasks requiring comparative reasoning or estimation under uncertainty. This raises questions about the consistency of ViT-based interpretation in visualization systems, especially when human-like perceptual fidelity is required. 
Our findings on task-specific divergences — for example, ViTs treating curvature and area as easier than humans, or performing worse in comparative judgments like position-length and point cloud perception — suggest that such perceptual misalignments could have practical consequences. In particular, they may affect downstream applications such as automated chart summarization and perceptually optimized visualization generation, where alignment with human perception is critical for producing effective and trustworthy outcomes.

\subsection{CNNs vs. ViTs}
The previous findings by Haehn et al.'s study on CNNs~\cite{haehn2018evaluating} emphasize the superior regression capabilities of CNNs over humans in elementary perceptual tasks. Our study on ViTs aligns with these findings, yet it falls short of outperforming CNNs by a margin of error of 6$\%$, as demonstrated in Table~\ref{tab:trained_models_ele}. However, upon comparing the ranking of CNNs and ViTs with humans across various perceptual tasks, ViTs exhibit closer alignment with human ranking than CNNs as shown in Fig.~\ref{fig:figure_ranking}. ViTs share similar ranking as humans for position non-aligned scale (rank:2), direction (rank:3) and shading (rank:6). 
Although ViTs are capable of interpolating between training datapoints, our findings indicate, that their ability to generalize across broader parameter variations is limited, similar to CNNs. In particular, the Swin Transformer showed decreasing accuracy as the complexity of parameterization increased, consistent with observations from CNNs in Haehn et al.~\cite{haehn2018evaluating}. Moreover, similar to the VGG19 model reported Haehn's study, the Swin Transformer struggled with generalization to transformations not seen during training, such as changes in width or spatial translations. These results suggest that, despite architectural differences, ViTs share similar limitations with CNNs when it comes to handling variability in visual encoding.
In regression analyses across other tasks such as position-angle, position-length, bars, framed rectangles, and point cloud perception, CNNs perform better than ViTs, with average errors ranging from 2.93 to 4.21, 3.97 to 4.72, 1.93 to 4.75, and 3.40 to 6.37, respectively. This behaviour might be attributed to the absence of a local receptive field and the data-intensive nature of ViTs.

Further, in our investigation of various ViT architectures, we observed that the Swin transformer demonstrated superior performance compared to others, including CvT, despite CvTs draw inspiration from CNNs. This discrepancy in performance could perhaps be attributed to Swin's hierarchical design and its utilization of spatial and sliding window mechanisms, features that share similarities with CNNs.
Thus, we believe that their strength lies in effectively combining both local and global spatial information and utilizing an extensive array of trainable parameters compared to CNNs. However, in general, ViTs encounter difficulties when attempting to interpret information from low-level visual tasks and are therefore not reliable for decision-making in the field of visualization.

\subsection{Ablation Studies}
To ensure that the reported findings are not bound to a specific parameter set, we have conducted ablation studies in order to analyze the behavior of ViTs under varying parameters. Tables~\ref{tab:pre_trained_models_ele} (bottom) and ~\ref{tab:ablation_study} show the results of our ablation settings.

\noindent\textbf{Image resolution.} ViTs are typically optimized to process images at a resolution of $224 \times 224$. Rather than merely resizing the images to fit this requirement, we generated a separate dataset with this resolution. However, this adjustment did not yield improved performance as shown in Table~\ref{tab:ablation_study}.

\noindent\textbf{Patch size.} Besides image resolution, patch size has a great impact on the outcome of ViTs. Nevertheless, in the past, ViTs have shown strong results in image classification with relatively large patches, we conducted an ablation study on vViTs using smaller patches, particularly of size 8, to evaluate their impact on low-level perceptual tasks. However, as shown in Table~\ref{tab:ablation_study} (right), this change did not lead to improved performance. Consequently, it is evident that the optimal balance between computational efficiency and perceptual accuracy is not solely determined by patch size. This suggests that simply reducing patch size does not guarantee better perceptual alignment in low-level tasks, and that architectural or training adaptations may be necessary to leverage finer granularity effectively.

\noindent\textbf{Training data size.} While CNNs traditionally rely on datasets of a certain size to maintain consistency, ViTs excel with larger datasets. Therefore, we opted to enhance training by utilizing datasets four times larger than the original, aligning with the capacity of transformers to leverage extensive data for better performance. Despite their data excessive need, we did not find the performance enhanced when increasing training data size as shown in Table~\ref{tab:ablation_study}.

\noindent\textbf{Pretraining.} In our ablation study, we utilized pretrained transformers as the foundation for finetuning experiments. Leveraging pretrained models allows our experiments to start from a highly informative point. For this purpose, we leveraged ViTs pre-trained on natural images sourced from the widely recognized ImageNet dataset~\cite{NIPS2012_c399862d}. Our findings reveal two noteworthy observations. Firstly, our best-performing network, Swin, exhibited superior performance compared to using pretrained weights.
However, Table~\ref{tab:pre_trained_models_ele} illustrates the overall average error on elementary perceptual tasks, which was lower when utilizing pretrained networks trained on the ImageNet dataset, with CvT achieving the lowest error rate. These findings suggest that while pretraining can enhance perceptual accuracy, its effectiveness depends on the model's structure and its compatibility with the target task.

\section{Conclusions and Future Work}
Our study sheds light on the perceptual capabilities of ViTs in comparison to both humans and CNNs. While ViTs exhibit superior performance compared to CNNs in general vision tasks, they still fall short of human performance. This raises concerns about their efficacy in data visualization, especially when requiring performance in low-level visual tasks. This 
raises concerns about their reliability in tasks that demand perceptual accuracy and interpretability.
We believe that our work highlights the need for further research, and underlines the need for redesigning applications to better leverage the strengths of ViTs in data visualization tasks.

In our study, we have focused our investigation on three specific types of architectures. However, there are promising avenues for future research 
that could provide deeper insights into their perceptual limitations and strengths. For example, exploring larger ViT models may reveal how model scale affects alignment with human perception.
Additionally, further analysis of the CvT architecture is warranted, as it combines elements from both CNNs and ViTs and has demonstrated improved performance when pretrained with natural images. Moreover, while our work has primarily utilized open-source ViT architectures, 
future work could also explore closed-source multimodal models like GPT-4o, which have shown potential in high-level visual reasoning.

Integrating these diverse models and delving deeper into their interactions could yield valuable insights applicable across a wide spectrum of visualization applications. All these efforts have the potential to further enhance our understanding of ViTs' visual perception capabilities.
Ultimately, our findings lay the groundwork for more perceptually aligned model development and evaluation in the graphics and visualization community.

%---------------------------------------------Paper Ends-------------------------------------------------%

%\begin{comment}
\section*{Acknowledgements}   
%\end{comment}
%-----------------------------------------------------------
%\section*{References}
\bibliographystyle{cag-num-names}
\bibliography{abbreviations,bibliography}
%-----------------------------------------------------------

\end{document}